\documentclass[conference]{IEEEtran}

\ifCLASSINFOpdf
   \usepackage[pdftex]{graphicx}
   \graphicspath{{figures/}}
   \DeclareGraphicsExtensions{.pdf,.jpeg,.png}
\else
   \usepackage[dvips]{graphicx}
   \graphicspath{{figures/}}
   \DeclareGraphicsExtensions{.eps}
\fi

\usepackage{cite}

\usepackage{amsmath}
\usepackage{array}

\ifCLASSOPTIONcompsoc
  \usepackage[caption=false,font=normalsize,labelfont=sf,textfont=sf]{subfig}
\else
  \usepackage[caption=false,font=footnotesize]{subfig}
\fi
\usepackage{float}

\begin{document}

\title{An Improved LPTC Neural Model for Background Motion Direction Estimation}

\author{\IEEEauthorblockN{Hongxin Wang}
\IEEEauthorblockA{School of Computer Science\\
University of Lincoln\\
Lincoln, LN6 7TS, UK \\
13488337@students.lincoln.ac.uk}
\and
\IEEEauthorblockN{Jigen Peng}
\IEEEauthorblockA{School of Mathematics and Statistics\\
 Xi'an Jiaotong University \\
Xi'an, Shaanxi, 710049, P.R. China\\
jgpeng@mail.xjtu.edu.cn}
\and
\IEEEauthorblockN{Shigang Yue}
\IEEEauthorblockA{School of Computer Science\\
University of Lincoln \\
Lincoln, LN6 7TS, UK \\
syue@lincoln.ac.uk}}

\maketitle

\begin{abstract}
A class of specialized neurons, called lobula plate tangential cells (LPTCs) has been shown to respond strongly to wide-field motion. The classic model, elementary motion detector (EMD) and its improved model, two-quadrant detector (TQD) have been proposed to simulate LPTCs. Although EMD and TQD can percept background motion, their outputs are so cluttered that it is difficult to discriminate actual motion direction of the background. In this paper, we propose a max operation mechanism to model a newly-found transmedullary neuron Tm9 whose physiological properties do not map onto EMD and TQD. This proposed max operation mechanism is able to improve the detection performance of TQD in cluttered background by filtering out irrelevant motion signals. We will demonstrate the functionality of this proposed mechanism in wide-field motion perception.
\end{abstract}

\IEEEpeerreviewmaketitle


\section{Introduction}
When flies search for and track prey or conspecifics, their own motion generates displacement of the visual surroundings, inducing wide-field background motion across the retina \cite{fox2014figure-2}. A class of specialized neurons, called lobula plate tangential cells (LPTCs), has been shown to respond strongly to wide-field motion. LPTCs can be broadly divided into a vertical system (VS) and a horizontal system (HS), which signal wide-field motion in vertical and horizontal directions, respectively \cite{lee2015spatio}. 

The classic correlation model, elementary motion detector (EMD) \cite{hassenstein1956systemtheoretische} and its improved model, two-quadrant detector (TQD) \cite{eichner2011internal,fu2017modeling} have been proposed to simulate LPTC neurons. These two models show strong responses to wide-field motion and have a clear mapping onto neural circuits of fly visual system. Although EMD and TQD are able to detect background motion, detection performances of these two models are always unsatisfying, especially in  cluttered environment. Due to indiscriminate signal correlation, both EMD and TQD always have four outputs which do not show much differences in the strength, representing lobula plate tangential cells' (LPTCs) neural responses along four cardinal directions (up, down, left, right). In some cases, model outputs along actual motion direction is even weaker than model outputs along other directions. Therefore, the actual direction of target motion cannot be determined by simply comparing the strengths of model outputs along four cardinal directions.

Recently, biologists have identified a transmedullary neuron, Tm9 whose physiological properties do not map onto classic EMD and TQD models, but is required for motion perception \cite{fisher2015class}. Further research indicates that the receptive field of Tm9 is much larger than that of its downstream neurons T5. Besides, signals from multi columns are converging at the level of Tm9. These findings are surprising especially when we consider that only signals from two adjacent photoreceptors are needed for motion computation in both EMD and TQD models. Based on these findings, we infer that Tm9 neurons are able to inform downstream neurons about local points in a wide receptive field. This property of Tm9 may help flies effectively avoid confusion caused by incorrect signal correlation while perceiving wide-field motion.

In this paper, we propose a max operation mechanism to simulate Tm9 neurons in order to improve the detection performance of TQD in cluttered background. This mechanism which acts on signals after ON-OFF channel separation of TQD is able to inform downstream neuron T4 and T5 about spatial maximum of ON and OFF signals in a local neighborhood. These local maximal signals are then temporally-delayed and integrated using the same method with classic TQD model. In the following paper, we will present modeling details of the improved TQD model, meanwhile demonstrating that the improved TQD model is able to overcome shortages of classic TQD model.

\section{Modeling}
\begin{figure}[!t]
\centering
\includegraphics[width=0.35\textwidth]{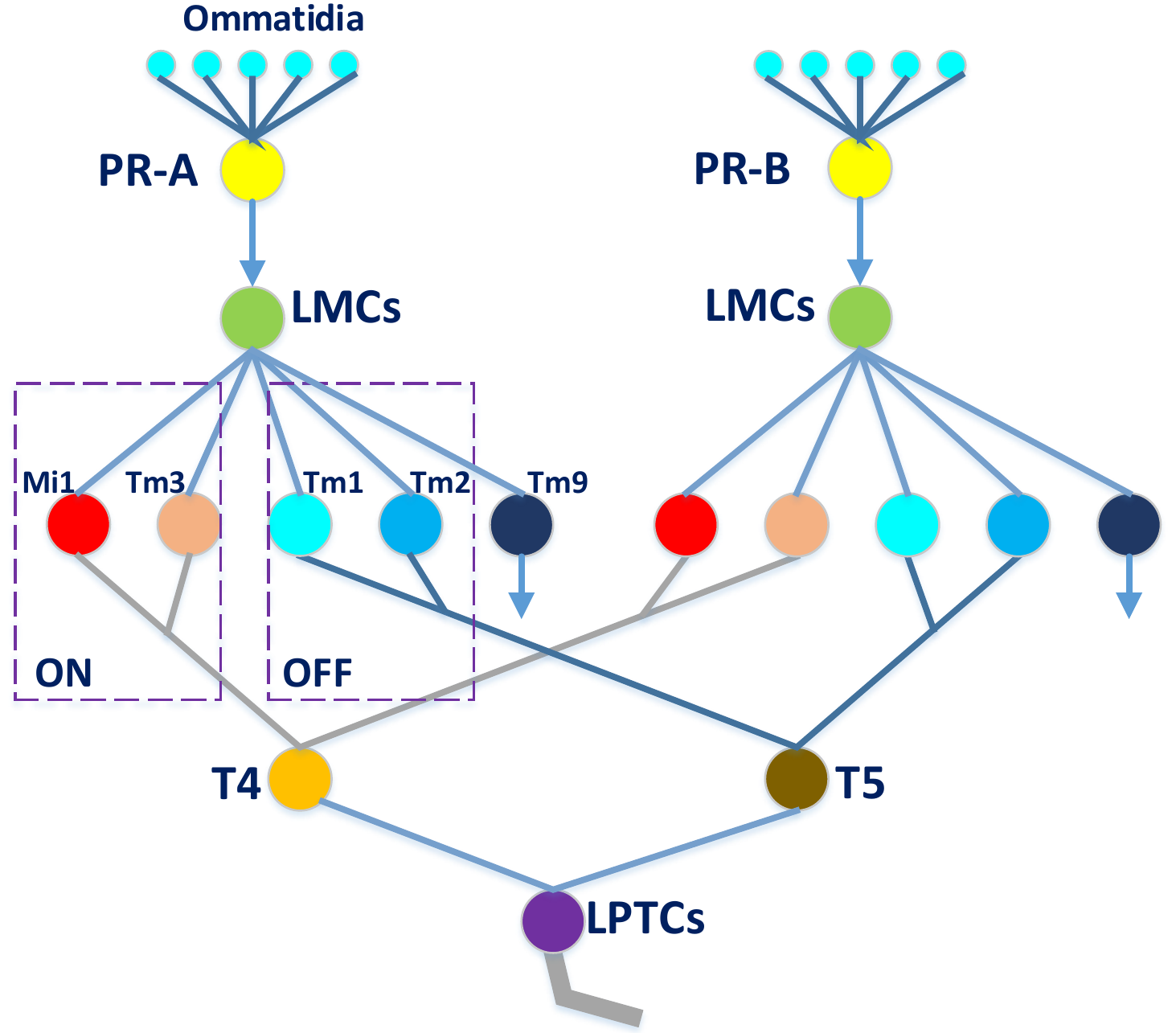}
\caption{The schematic of the improved TQD model, Each colorized disk denotes a neuron. PR-A, PR-B, LMCs, LPTCs are the abbreviation of photoreceptor A, Photoreceptor B, Large Monopolar Cells, Lobula Plate Tangential Cells, respectively.}
\label{Schematic-of-TQD}
\end{figure}

Fig. \ref{Schematic-of-TQD} shows the schematic of the improved TQD model. For showing the difference between classic TQD and the improved TQD model, the connection between Tm9 and T4, T5 neurons was not plotted in Fig. \ref{Schematic-of-TQD}. We introduce the improved TQD model layer by layer in the following paper.

\subsection{Retina Layer}
In order to simulate the signal processing of photoreceptors, we start by representing visual stimuli as varying luminance values, noted $I(x,y,t) \in \mathbf{R}$, where $x,y$ and $t$ are spatial and temporal field positions. Then, the functionality of photoreceptors is described by the following equation,
\begin{equation}
L(x,y,t) =  \iint I(u,v,t)G_{\sigma_1}(x-u,y-v)dudv
\label{Gaussian-Blur}
\end{equation}
where $L(x,y,t)$ is the output of photoreceptors. Particularly, for a given spatial position $(x_0,y_0)$, $I(x_0,y_0,t)$ represents luminance information received by the ommatidia located in spatial position $(x_0,y_0)$ at time $t$ while $L(x_0,y_0,t)$ denotes the output of the photoreceptor located in position $(x_0,y_0)$ at time $t$. $G_{\sigma_1}(x,y)$ is a Gaussian function, defined as
\begin{equation}
G_{\sigma_1}(x,y)= \frac{1}{2\pi\sigma_1^2}\exp(-\frac{x^2+y^2}{2\sigma_1^2}).
\label{Gauss-blur-Kernel}
\end{equation}

\subsection{Lamina Layer}
Photoreceptors synapse on large monopolar cells (LMCs) located in lamina layer which are able to remove redundancy contained in input signals ($L(x,y,t)$) and maximize information about illumination change. Here, we implement a temporal contrast detector on input signal ($L(x,y,t)$) so as to simulate neural responses of LMCs. That is,
\begin{align}
P(x,y,t) &= \int L(x,y,s)H(t-s) ds \\
H(t) &= \Gamma_{n_1,\tau_1}(t) - \Gamma_{n_2,\tau_2}(t)
\label{High-Pass-Filter-Kernel}
\end{align}
where $P(x,y,t)$ and $\Gamma_{n,\tau}(t)$ are the output of LMCs, Gamma function, respectively. $\Gamma_{n,\tau}(t)$ is defined as
\begin{equation}
\Gamma_{n,\tau}(t) = (nt)^n \frac{\exp(-nt/\tau)}{(n-1)!\tau^{n+1}}.
\end{equation}

Before LMCs relay processed signals ($P(x,y,t)$) to medulla layer, LMCs receive lateral inhibition from adjacent neurons. In accord with classic lateral inhibition mechanism, we convolve signal $P(x,y,t)$ with a inhibition kernel $W_1(x,y,t)$. That is,
\begin{equation}
P_I(x,y,t) = \iiint P(u,v,s)W_1(x-u,y-v,t-s) du dv ds
\label{First-Order-Lateral-Inhibition-Kernel}
\end{equation}
where $P_I(x,y,t)$ is laterally inhibited signal and $W_1(x,y,t)$ is defined using the following equations,
\begin{equation}
\begin{split}
W_1(x,y,t) = &W_S^{Pos}(x,y)W_T^{Pos}(t) \\ & \hspace{2em}...+W_S^{Neg}(x,y)W_T^{Neg}(t).
\end{split}
\end{equation}

In this paper, we set $W_S^{Pos}(x,y)$, $W_S^{Neg}(x,y)$, $W_T^{Pos}(t)$, $W_T^{Neg}(t)$ as
\begin{align}
W_S^{Pos} &= [G_{\sigma_2}(x,y) - G_{\sigma_3}(x,y)]^+ \\
W_S^{Neg} &= [G_{\sigma_2}(x,y) - G_{\sigma_3}(x,y)]^- ,  \ \sigma_3 = 2*\sigma_2\\
W_T^{Pos} &= \frac{1}{\alpha_1}\exp(-\frac{t}{\alpha_1})  \\
W_T^{Neg} &= \frac{1}{\alpha_2}\exp(-\frac{t}{\alpha_2}),  \ \alpha_2 > \alpha_1 .
\end{align}
where $[x]^+, [x]^-$ denote $\max (x,0)$ and $\min (x,0)$, respectively. $G_{\sigma_2}(x,y)$ and $G_{\sigma_3}(x,y)$ are Gaussian functions.

\subsection{Medulla Layer}
\label{Subsection-Mdeulla}
Previous research identified two parallel pathways which selectively respond to brightness increments (ON pathway) and decrements (OFF pathway) in medulla layer \cite{behnia2014processing,joesch2010and}. These two parallel pathways are implemented by four intermediate neurons, i.e., Tm1, Tm2, Tm3 and Mi1. To be more precise, Mi1 and Tm3 constitute ON pathway whereas Tm1 and Tm2 form OFF pathway, shown in Fig. \ref{Schematic-of-TQD}. Besides, compared to the output of Tm3, the output of Mi1 is temporally delayed. Similarly, the output of Tm1 has a temporal delay compared to that of Tm2.

Based on these biological findings, TQD which is the improved model of EMD firstly separate laterally inhibited signal $P_I(x,y,t)$ into ON and OFF channels, shown in Fig. \ref{Schematic-of-TQD}. That is, 
\begin{align}
S^{ON}(x,y,t) &= \frac{(|P_I(x,y,t)|+ P_I(x,y,t))}{2}  \\
S^{OFF}(x,y,t) &= \frac{(|P_I(x,y,t)|-P_I(x,y,t))}{2}
\end{align}
where $S^{ON}$ and $S^{OFF}$ denote the output of Tm3 and Tm2, respectively.

Due to small temporal delay exists between the outputs of Mi1 and Tm3 (the outputs of Tm1 and Tm2), signal $S^{ON}$ ($S^{OFF}$) is convolved with a Gamma function so as to obtain the delayed signal  $S_D^{ON}$ ($S_D^{OFF}$). This process can be described by the following equations,
\begin{align}
S_D^{ON}(x,y,t) &= \int S^{ON}(x,y,s)\Gamma_{n_3,\tau_3}(t-s) ds \label{ON-Channel-Delay} \\
S_D^{OFF}(x,y,t) &= \int S^{OFF}(x,y,s)\Gamma_{n_3,\tau_3}(t-s) ds
\label{OFF-Channel-Delay}
\end{align}
where $S_D^{ON}$ and $S_D^{OFF}$ are time-delayed signals, corresponding to the output of Mi1 and Tm1, respectively.

However, recent research has identified a transmedullary neuron Tm9 whose physiological properties do not map onto classic TQD model but which is required for motion perception \cite{fisher2015class}. Compared to other neurons, such as Tm1, Tm2, Tm3 and Mi1, Tm9 has much larger receptive field which conflicts with the view that downstream neurons, like T4 and T5, only require signals from two neighboring photoreceptors in a relatively small receptive field. Because Tm9 has a larger receptive field, signals from multi columns (or photoreceptors) can be integrated in Tm9. Obviously, this functionality cannot be accomplished by other neurons, like Tm1, Tm2, Tm3 and Mi1, which always only receive signal from a single column (or photoreceptor). Based on these biological findings, we assume that Tm9 is able to compare the strength of signals received from multi columns and find a local maximum. In order to account for properties of Tm9, we propose a max operation mechanism acting on ON and OFF signals ($S^{ON}$ and $S^{OFF}$). This max operation mechanism can be described by the following equations,
\begin{equation}
\tilde{S}^{ON}(x_0,y_0,t) =
\begin{cases}
0 & \text{if} \ \text{flag}^{ON} = 0\\
S^{ON}(x_0,y_0,t) & \text{if} \ \text{flag}^{ON} = 1
\end{cases}
\end{equation}

\begin{equation}
\tilde{S}^{OFF}(x_0,y_0,t) =
\begin{cases}
0 & \text{if} \  \ \text{flag}^{OFF} = 0\\
S^{OFF}(x_0,y_0,t) & \text{if} \ \  \text{flag}^{OFF} = 1
\end{cases}
\end{equation}
where $\text{flag}^{ON}$ and $\text{flag}^{OFF}$ are defined by the following equations,
\begin{equation}
\text{flag}^{ON}=
\begin{cases}
0 & \text{if} \ S^{ON}(x_0,y_0,t) \neq \max_{(x,y) \in \Omega_{(x_0,y_0)}} S^{ON}\\\\
1 & \text{if}\ S^{ON}(x_0,y_0,t) = \max_{(x,y) \in \Omega_{(x_0,y_0)}} S^{ON}
\end{cases}
\end{equation}

\begin{equation}
\text{flag}^{OFF}=
\begin{cases}
0 & \text{if} \ S^{OFF}(x_0,y_0,t) \neq \max_{(x,y) \in \Omega_{(x_0,y_0)}} S^{OFF}\\\\
1 & \text{if}\ S^{OFF}(x_0,y_0,t) = \max_{(x,y) \in \Omega_{(x_0,y_0)}} S^{OFF}
\end{cases}
\end{equation}
where $\Omega_{(x_0,y_0)}$ is a local neighborhood centered at $(x_0,y_0)$.

\begin{figure}[t!]
	\centering
	\subfloat[Original Signal]{\includegraphics[width=0.2\textwidth]{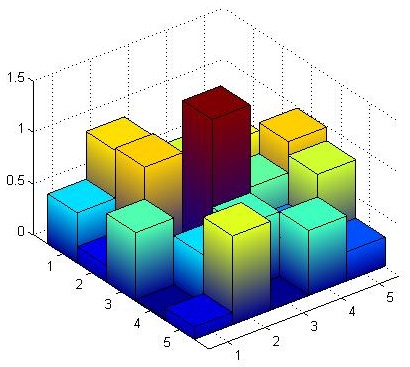}
		\label{Example-Max-Operation-Orignal}}
	\hfil
	\subfloat[Signal After Max Operation]{\includegraphics[width=0.2\textwidth]{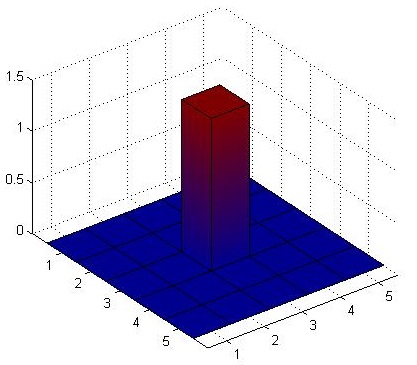}
		\label{Example-Max-Operation-Processed}}
	\caption{Schematic illustration of the proposed max operation mechanism.}
	\label{Example-Max-Operation}
\end{figure}

In order to clearly illustrate this max operation mechanism, an example is shown in Fig. \ref{Example-Max-Operation}. As we can see from Fig. \ref{Example-Max-Operation}, $S^{ON}(3,3,t_0)$ (or $S^{OFF}(3,3,t_0)$) is the local maximum in the local neighborhood $\Omega_{(3,3)}$, so $S^{ON}(3,3,t_0)$ (or $S^{OFF}(3,3,t_0)$) is preserved after max operation. However, because other signals, for example $S^{ON}(1,1,t_0)$，$S^{ON}(1,2,t_0)$， $S^{ON}(2,2,t_0)$ and so on, are not local maximum in the local neighborhood, these signals are set as $0$ after max operation. Obviously, this max operation mechanism is able to decrease clutter and increase sparsity of input signals. Due to the increment of sparsity, incorrect signal correlation will be effectively avoided in signal-correlation step.

After max operation, signal $\tilde{S}^{ON}(x,y,t)$ and $\tilde{S}^{OFF}(x,y,t)$ are temporally delayed. This step is similar with time-delay operation shown in Eq.(\ref{ON-Channel-Delay}) and (\ref{OFF-Channel-Delay}).
\begin{align}
\tilde{S}_D^{ON}(x,y,t) &= \int \tilde{S}^{ON}(x,y,s)\Gamma_{n_3,\tau_3}(t-s) ds  \\
\tilde{S}_D^{OFF}(x,y,t) &= \int \tilde{S}^{OFF}(x,y,s)\Gamma_{n_3,\tau_3}(t-s) ds
\end{align}
where $\tilde{S}_D^{ON}(x,y,t)$ and $\tilde{S}_D^{OFF}(x,y,t)$ are temporally-delayed signals.

\subsection{Lobula Layer}
In lobula layer, various high-order neurons integrate signals relayed from ON and OFF pathways, then respond selectively to specific visual stimuli. For example, small target motion detectors (STMDs) show exquisite selectivity for small target motion. Lobula plate tangential cells (LPTCs) are sensitive to wide-field motion. Elementary small target motion detector (ESTMD) \cite{wang2016bio,wiederman2008model} and two-quadrant detector (TQD) \cite{eichner2011internal} have been proposed to simulate STMD and LPTC neurons, respectively. In this paper, we focus on LPTC neuron modeling for background motion direction detection.

T4 and T5 neurons which are pre-synaptic neurons of LPTCs integrate signals relayed from medulla layer. More precisely, T4 responds selectively to ON signals while T5 is specialized for OFF signals. Let $D^{T4}(x,y,t;\theta)$ and $D^{T5}(x,y,t;\theta)$ denote T4 and T5 neural responses at spatial-temporal coordinate $(x,y,t)$ along direction $\theta$, respectively. Then, in classic TQD model, $D^{T4}(x,y,t;\theta)$ and $D^{T5}(x,y,t;\theta)$ are given by the following equations,
\begin{align}
D^{T4}(x,y,t;\theta) &= S^{ON}(x,y,t)S^{ON}_D(x',y',t) \label{T4-Neural-Correlation-Pos} \\    
D^{T5}(x,y,t;\theta)& = S^{OFF}(x,y,t)S^{OFF}_D(x',y',t)
\label{T5-Neural-Correlation-Pos}
\end{align}
where $\theta \in \{0,\frac{\pi}{2},\pi,\frac{3\pi}{2}\}$.

Because T4 and T5 also receive signals from Tm9, T4 and T5 have corresponding outputs $\tilde{D}^{T4}$, $\tilde{D}^{T5}$,
\begin{align}
\tilde{D}^{T4}(x,y,t;\theta) &= \tilde{S}^{ON}(x,y,t)\tilde{S}^{ON}_D(x',y',t) \label{T4-Neural-Correlation-Pos-Tm9} \\    
\tilde{D}^{T5}(x,y,t;\theta)& = \tilde{S}^{OFF}(x,y,t)\tilde{S}^{OFF}_D(x',y',t)
\label{T5-Neural-Correlation-Pos-Tm9}
\end{align}
where $\theta \in \{0,\frac{\pi}{2},\pi,\frac{3\pi}{2}\}$. 

Lobula plate tangential cells (LPTCs) further integrate signals provided by T4 and T5 neurons.
 
\textbf{\textit{1) Classic TQD Model:}} For classic TQD model, LPTC output $F(x,y,t;\theta)$ is defined by the following equation, 
\begin{equation}
F(x,y,t;\theta) = D^{T4}(x,y,t;\theta) + D^{T5}(x,y,t;\theta).
\end{equation}

\textbf{\textit{2) Improved TQD Model:}} For improved TQD model, LPTC output $\tilde{F}(x,y,t;\theta)$ is defined by the following equation, 
\begin{equation}
\tilde{F}(x,y,t;\theta) = \tilde{D}^{T4}(x,y,t;\theta) + \tilde{D}^{T5}(x,y,t;\theta).
\end{equation}

The direction of background motion is determined by comparing the strength of LPTCs' neural responses at different directions. That is,
\begin{equation}
\Theta(t) = \arg \max_{\theta} \iint \tilde{F}(x,y,t;\theta) dx dy
\end{equation}
where $\Theta(t)$ denotes the motion direction of background at time $t$.

Numerous methods have been developed for optic flow estimation. Particularly, structure-tensor based methods construct the tensor for each pixel within its neighborhood, then convert the optic flow estimation problem to an eigenvalue analysis problem \cite{liu2003accurate,kratz2010tracking}. Compared to structure-based methods, the improved TQD model offers a totally different way to estimate background motion. Based on biological findings, TQD model detects background motion by correlating signals relayed from two photoreceptors. Although this correlation method is relatively simple, it reflects the signal processing mechanism and neural circuits of fly visual system.

\section{Experiment}

\begin{figure}[t]
	\centering
	\includegraphics[width=0.25\textwidth]{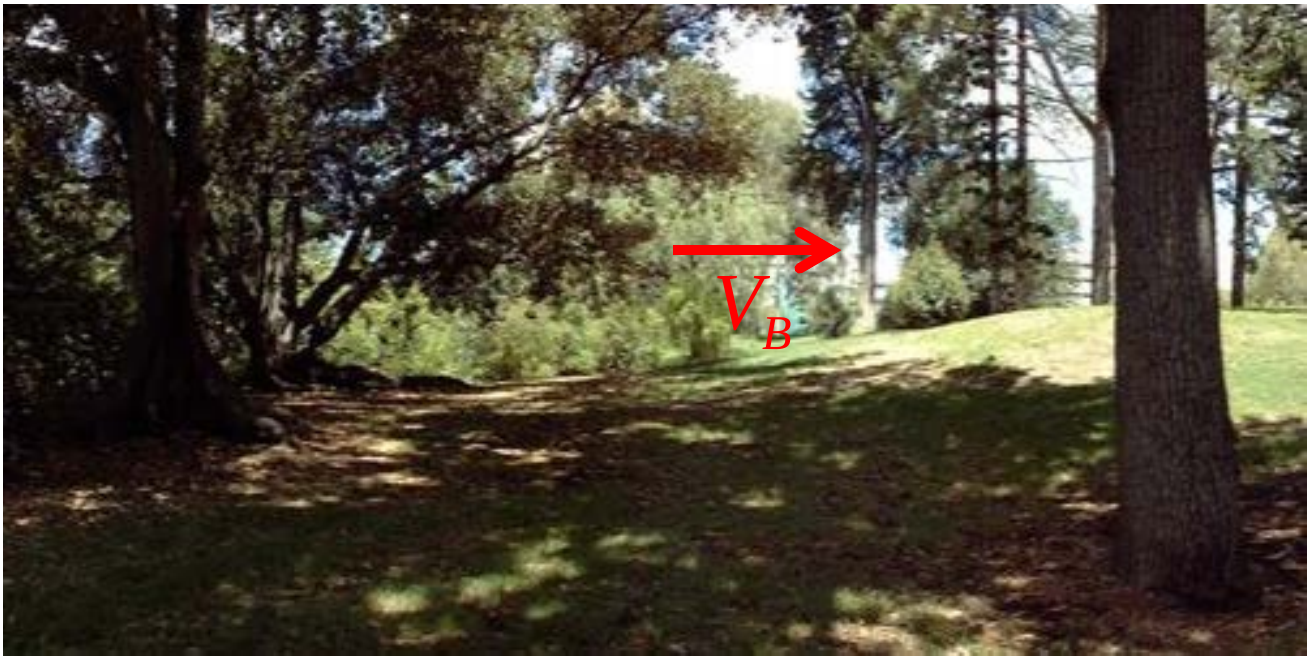}
	\caption{The $840$th frame of the first image sequence. The red arrow and $V_B$ denote motion direction and velocity of background, respectively.}
	\label{TestSet-1-Frame-840}
\end{figure}

\begin{figure}[t]
	\centering
	\subfloat[$F^{N}(x_0,y_0,t;\theta)$]{\includegraphics[width=0.2\textwidth]{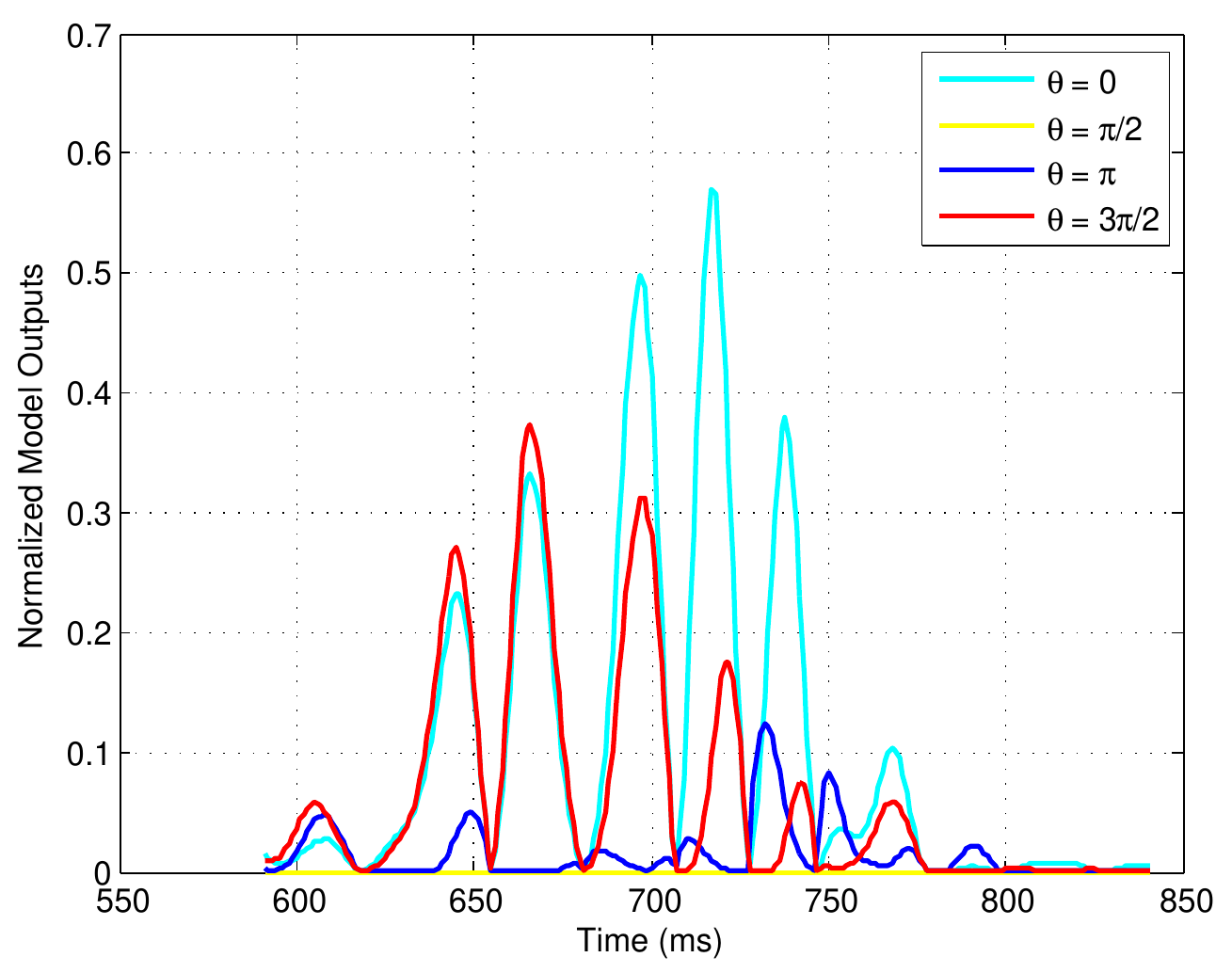}
		\label{Temporal-Field-x-113-y-80-NoMax}}
	\hfil
	\subfloat[$\tilde{F}^N(x_0,y_0,t;\theta)$]{\includegraphics[width=0.2\textwidth]{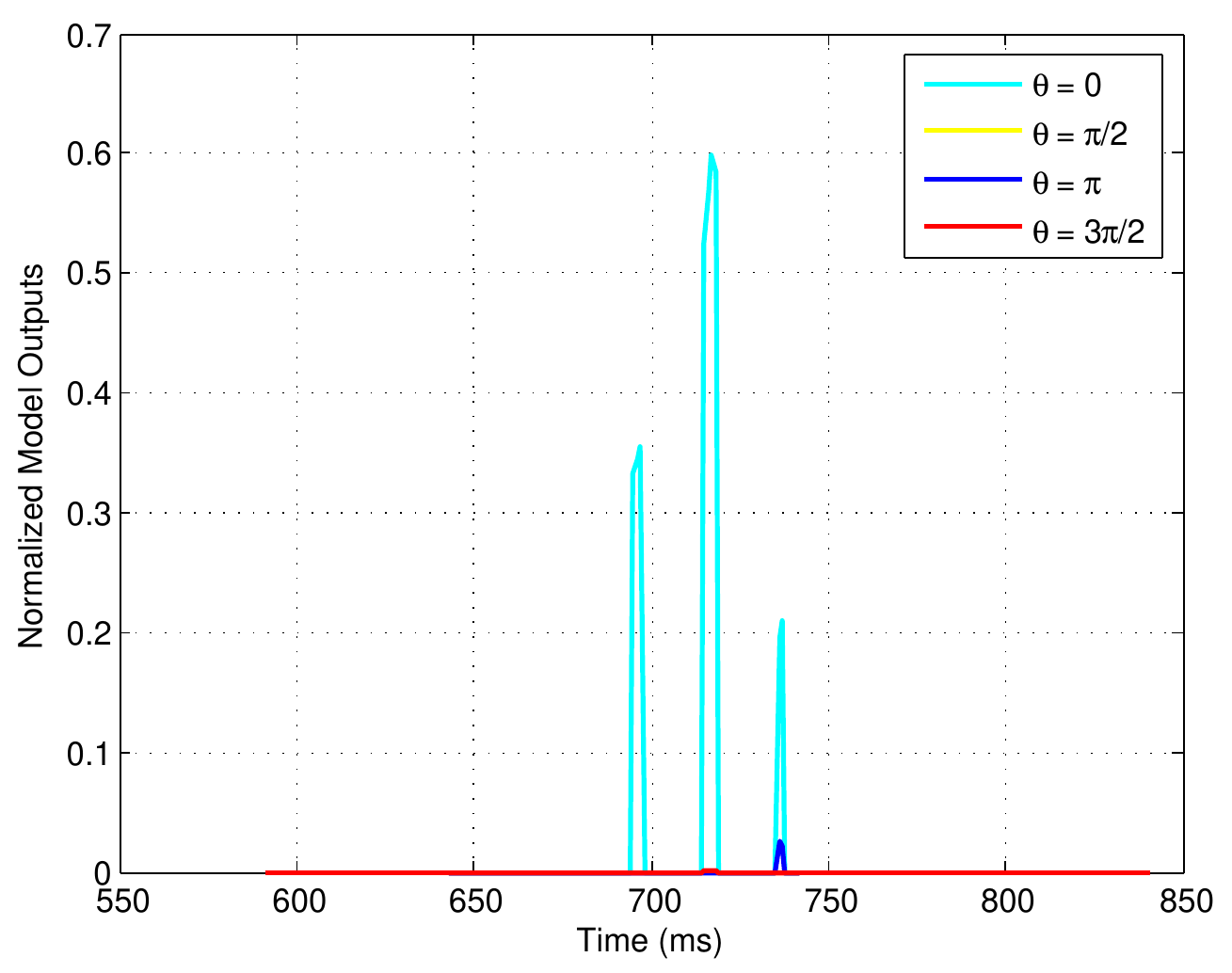}
		\label{Temporal-Field-x-113-y-80-Max}}
	\caption{The normalized model outputs of a spatial coordinate $(x_0,y_0)$ of the first image sequence during time period $[550,850]$ ms, i.e., $F^{N}(x_0,y_0,t;\theta)$, $\tilde{F}^{N}(x_0,y_0,t;\theta)$, $t\in [550,850]$ ms, $\theta \in \{0, \frac{\pi}{2}, \pi, \frac{3\pi}{2}\}$.}
	\label{Temporal-Field-x-113-y-80}
\end{figure}

In this section, three synthetic image sequences were used to evaluate detection performance of classic TQD and improved TQD models. The sampling frequencies of these three image sequences are all set as $1000$ Hz. Fig. \ref{TestSet-1-Frame-840} shows a frame of the first image sequence which is $500$ (in horizontal) by $250$ (in vertical) pixels. As we have mentioned before, background motion is caused by flies' ego-motion. Therefore, in this paper, background is in one of four cardinal direction motion (rightward, leftward, upward, downward) so as to simulate the displacement of flies' head. For example, in Fig. \ref{TestSet-1-Frame-840}, background is in rightward motion where red arrow and $V_B$ denote motion direction and velocity of background, respectively.

\begin{figure*}[!t]
	\centering
	\subfloat[$F(x,y,t;0) > \gamma$]{\includegraphics[width=0.2\textwidth]{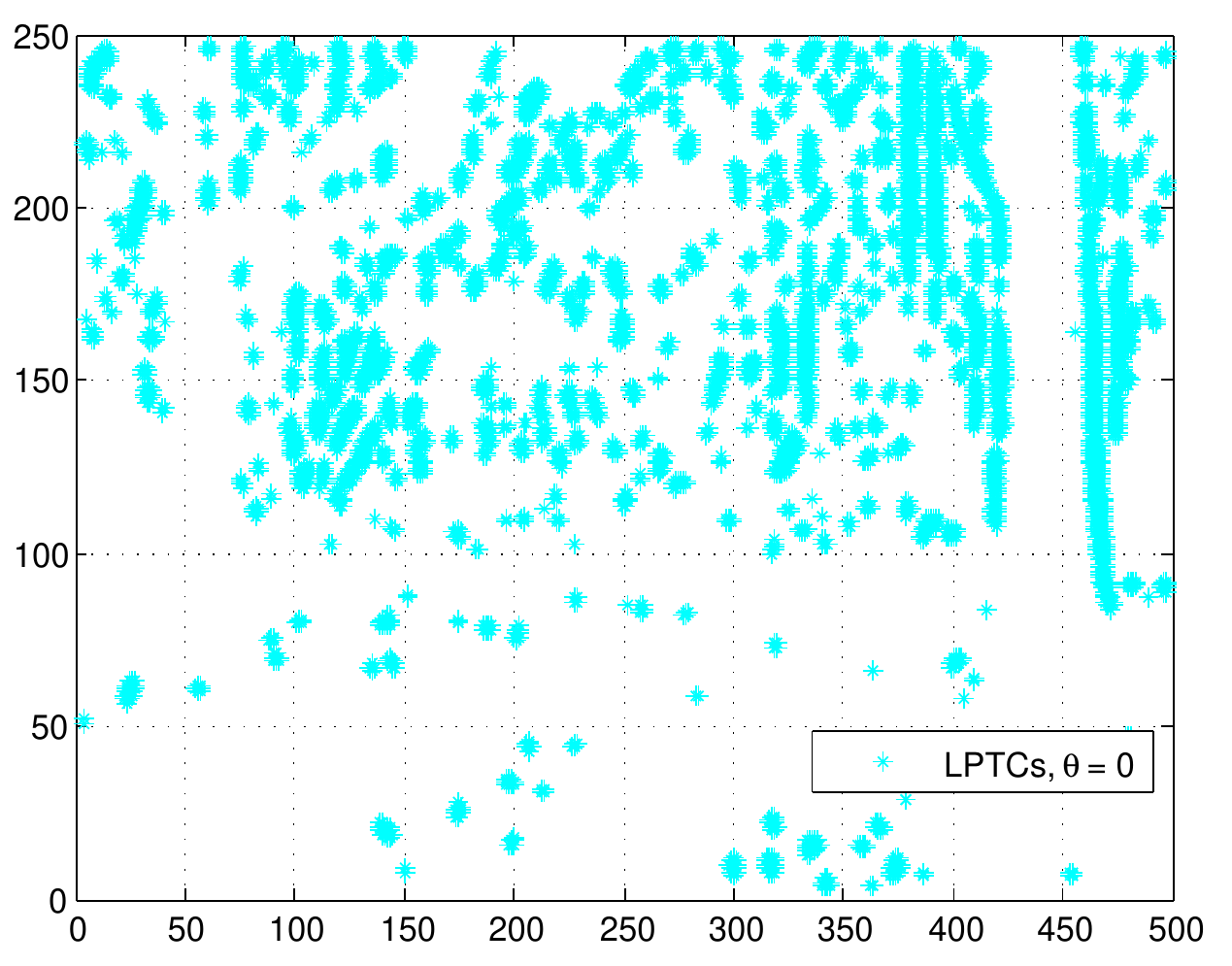}
		\label{NoMax-Projection-C1}}
	\hspace{1em}
	\subfloat[$F(x,y,t;\frac{\pi}{2}) > \gamma$]{\includegraphics[width=0.2\textwidth]{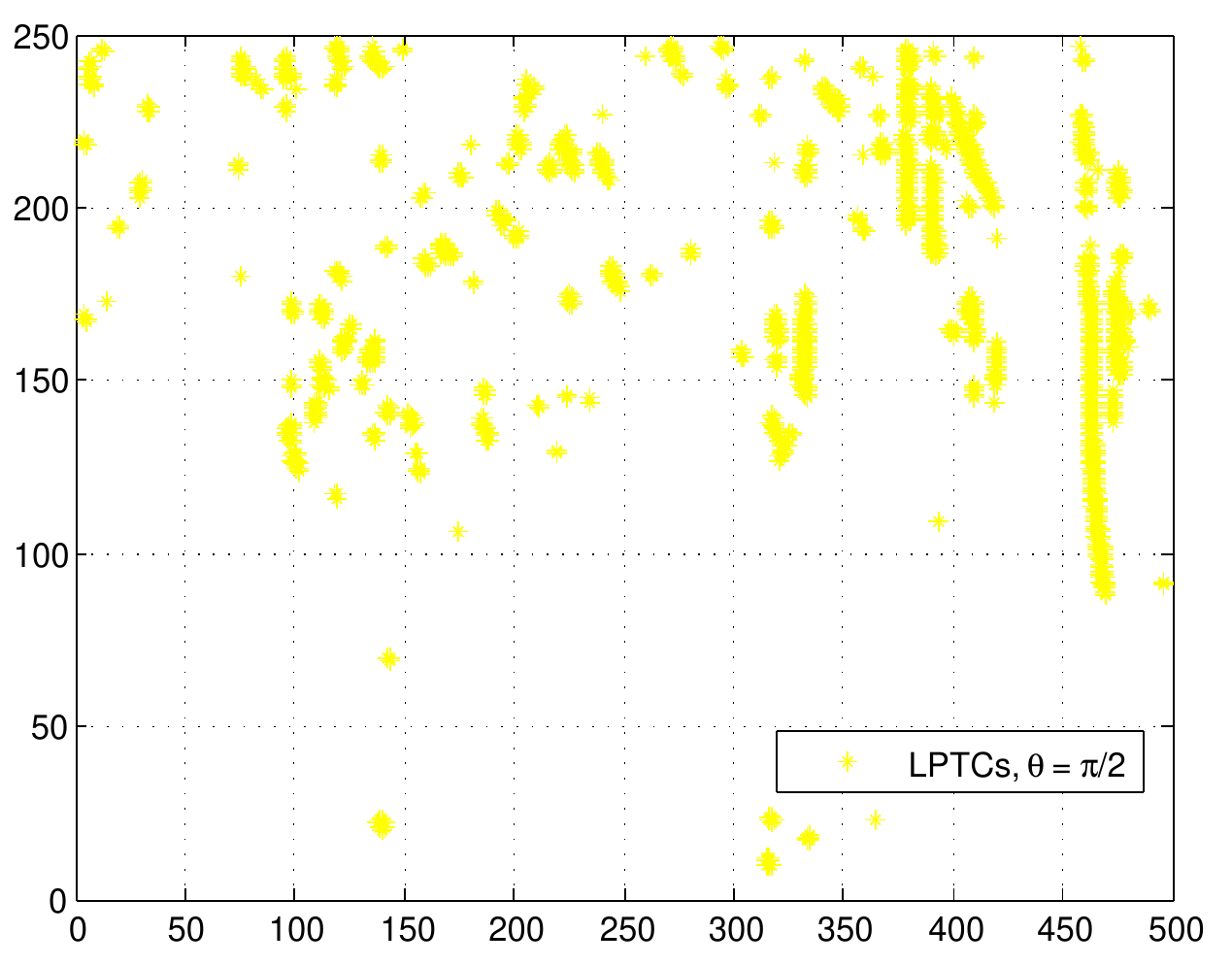}
		\label{NoMax-Projection-C2}}
	\hspace{1em}
	\subfloat[$F(x,y,t;\pi) > \gamma$]{\includegraphics[width=0.2\textwidth]{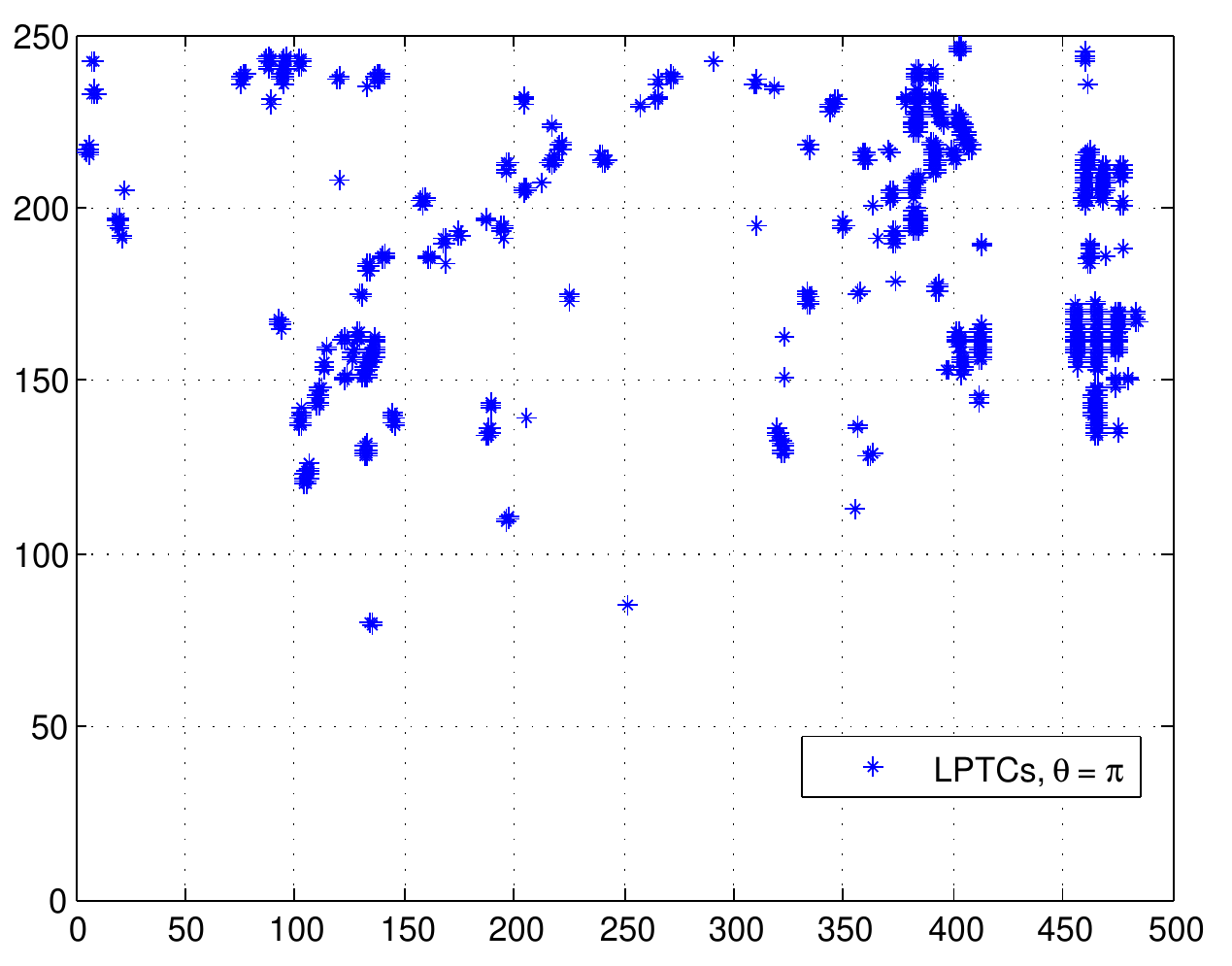}
		\label{NoMax-Projection-C3}}
	\hspace{1em}
	\subfloat[$F(x,y,t;\frac{3\pi}{2}) > \gamma$]{\includegraphics[width=0.2\textwidth]{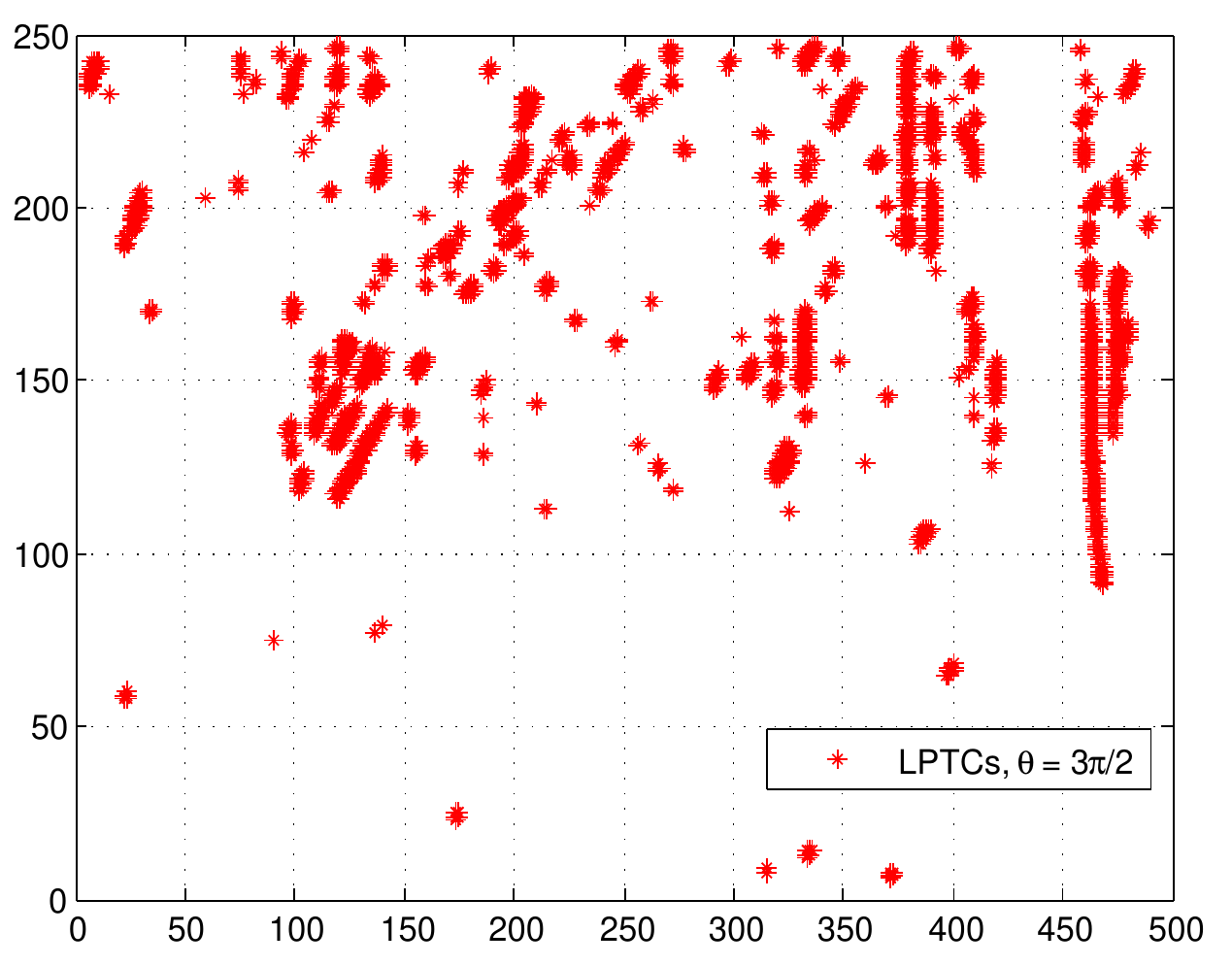}
		\label{NoMax-Projection-C4}}
	\caption{Projection results of normalized model outputs $F^{N}(x,y,t;\theta)$ of the first image sequence, where projection threshold is set as $0.05$ and $t$ is equal to $840$ ms.}
	\label{NoMax-Projection}
\end{figure*}

\begin{figure*}[!t]
	\centering
	\subfloat[$\tilde{F}(x,y,t;0) > \gamma$]{\includegraphics[width=0.2\textwidth]{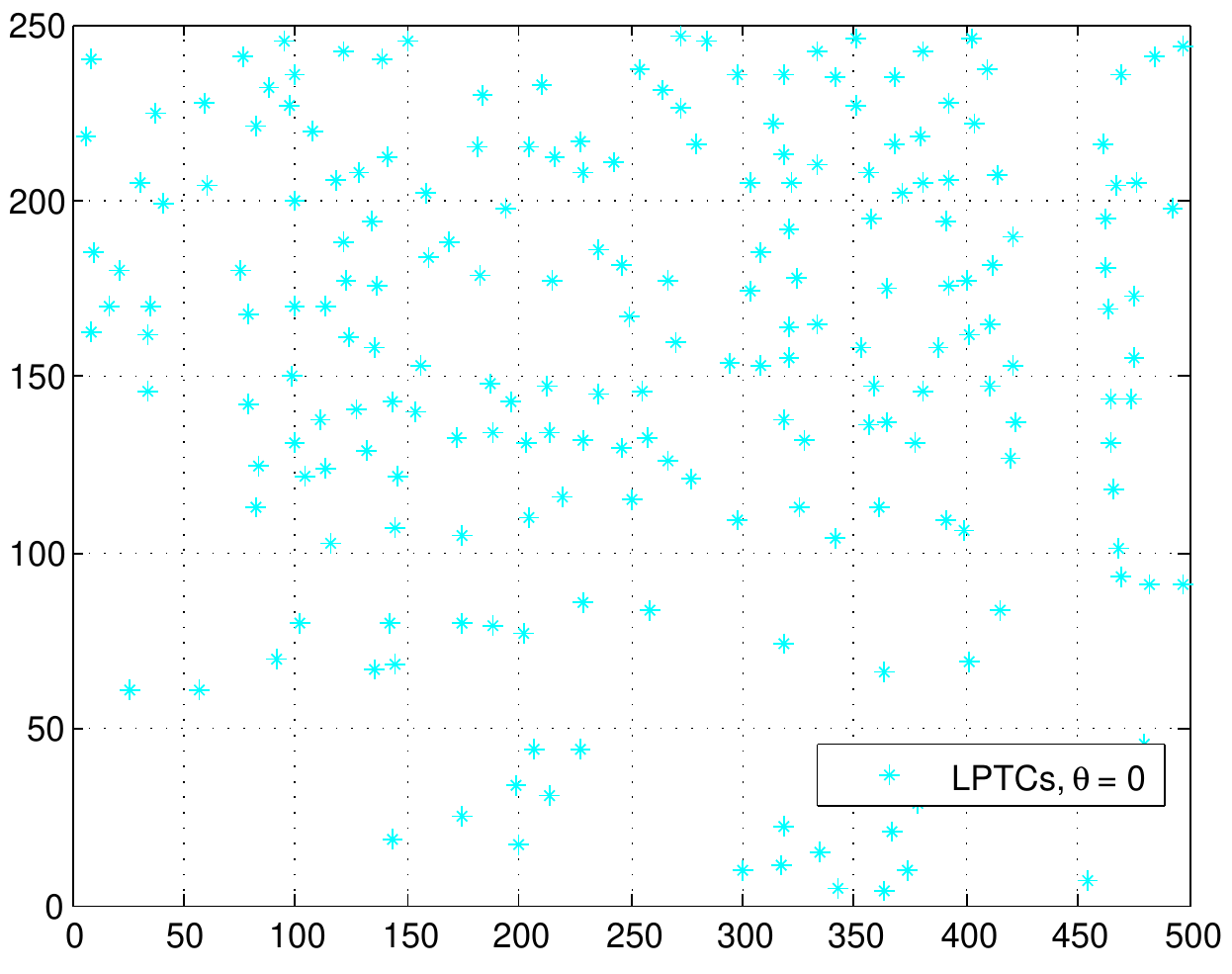}
		\label{Max-Projection-C1}}
	\hspace{1em}
	\subfloat[$\tilde{F}(x,y,t;\frac{\pi}{2}) > \gamma$]{\includegraphics[width=0.2\textwidth]{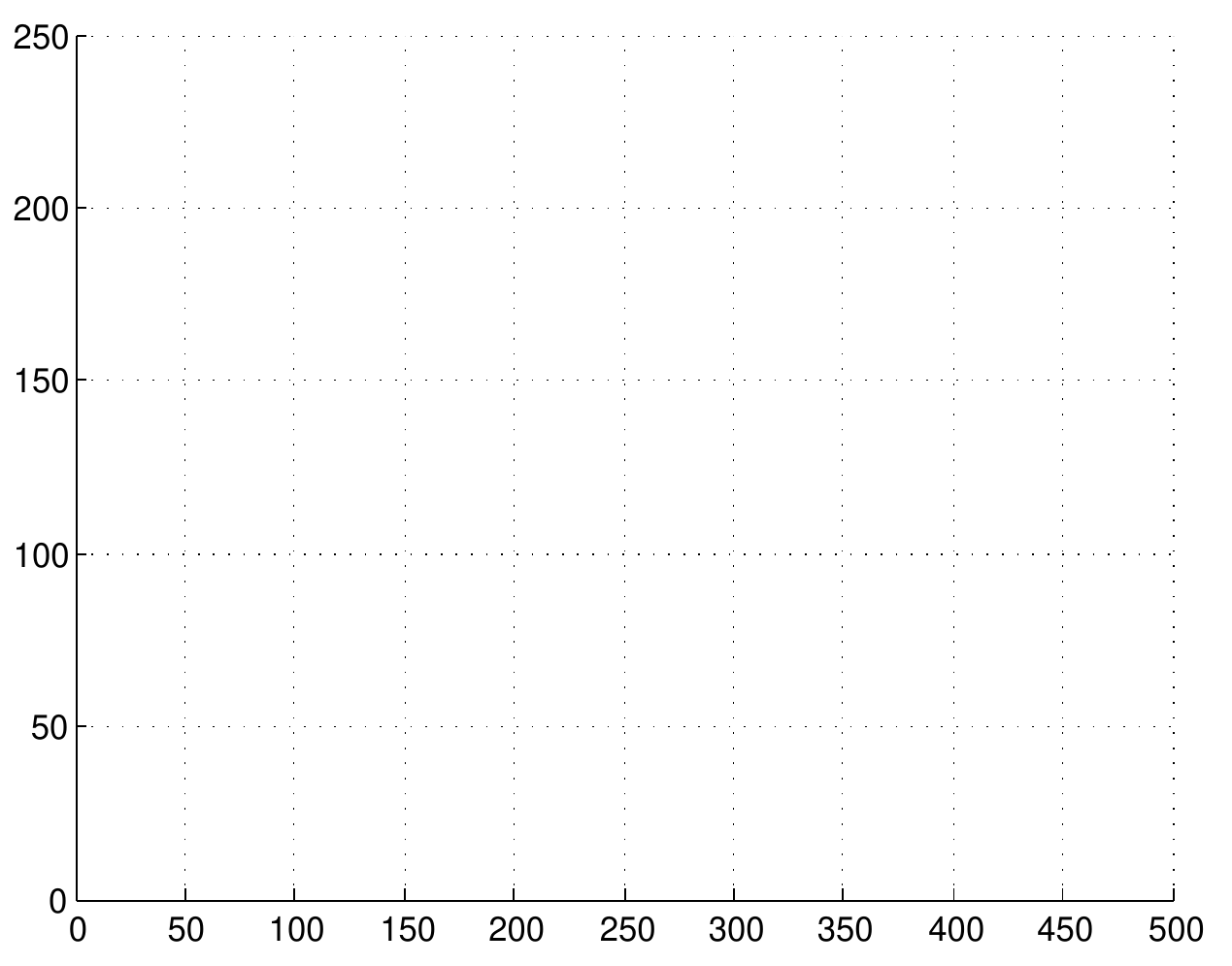}
		\label{Max-Projection-C2}}
	\hspace{1em}
	\subfloat[$\tilde{F}(x,y,t;\pi) > \gamma$]{\includegraphics[width=0.2\textwidth]{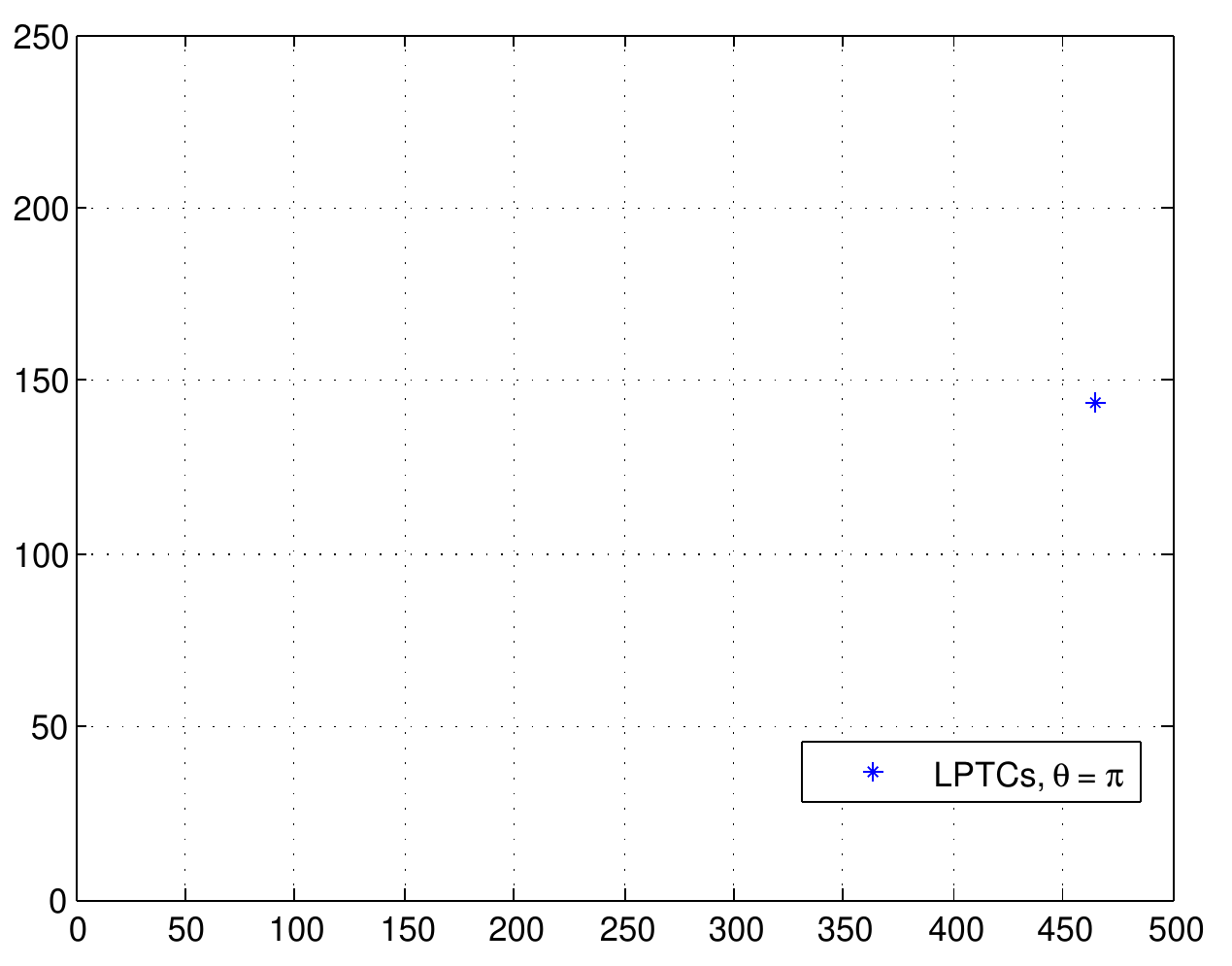}
		\label{Max-Projection-C3}}
	\hspace{1em}
	\subfloat[$\tilde{F}(x,y,t;\frac{3\pi}{2}) > \gamma$]{\includegraphics[width=0.2\textwidth]{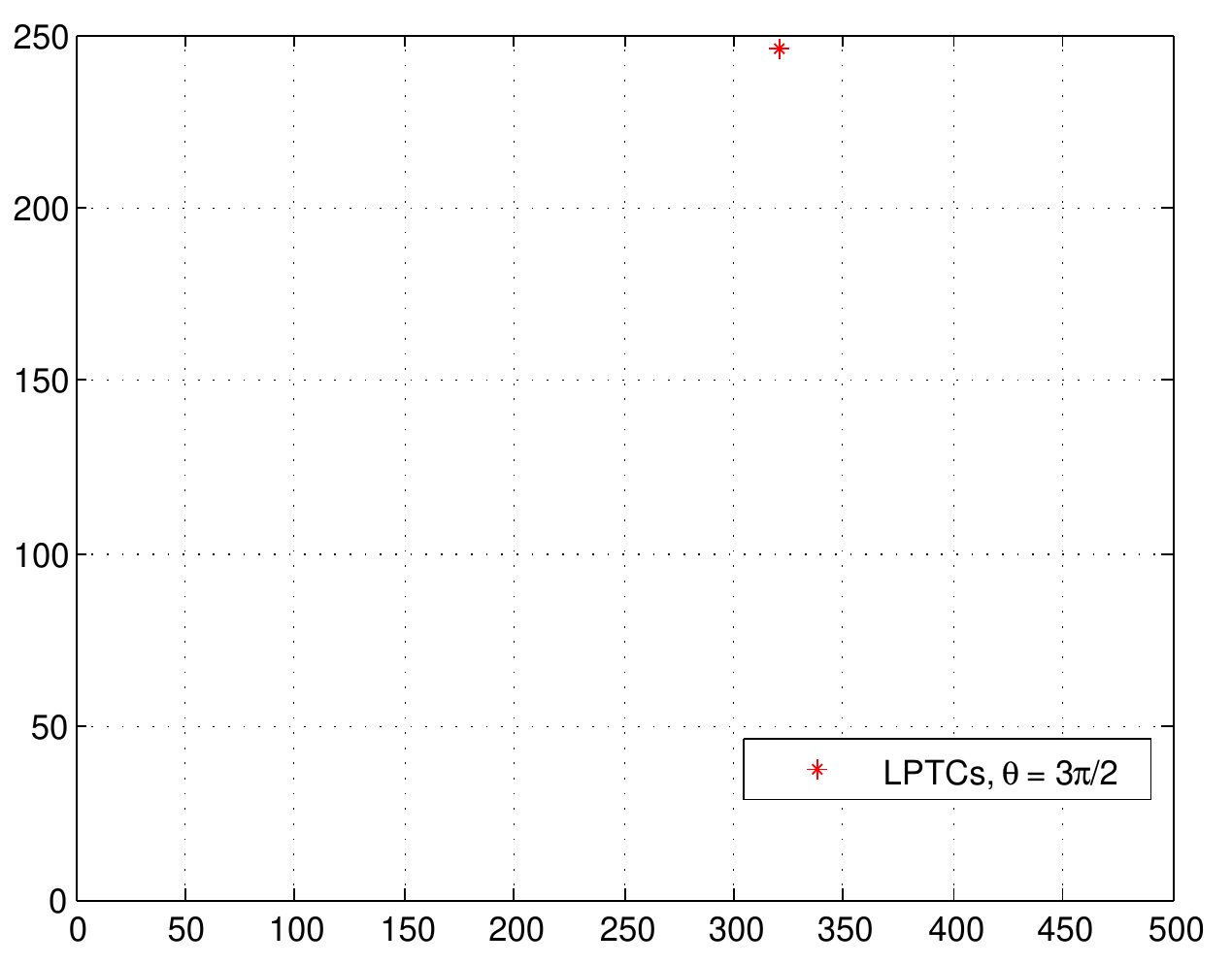}
		\label{Max-Projection-C4}}
	\caption{Projection results of normalized model outputs $\tilde{F}^{N}(x,y,t;\theta)$ of the first image sequence, where projection threshold is set as $0.05$ and $t$ is equal to $840$ ms.}
	\label{Max-Projection}
\end{figure*}

In order to compare classic TQD model output $F(x,y,t;\theta)$ and the improved TQD model output $\tilde{F}(x,y,t;\theta)$, $F(x,y,t;\theta)$ and $\tilde{F}(x,y,t;\theta)$ are firstly normalized. That is, 
\begin{align}
F^{N}(x,y,t;\theta) &= \frac{F(x,y,t;\theta)}{
	\max_{x,y,\theta}F(x,y,t;\theta)} \\
\tilde{F}^{N}(x,y,t;\theta) &= \frac{\tilde{F}(x,y,t;\theta)}{
	\max_{x,y,\theta}\tilde{F}(x,y,t;\theta)}.
\end{align}

Then, the normalized model outputs of a spatial coordinate $(x_0,y_0)$ during a time period $[550,850]$ ms, i.e., $F^{N}(x_0,y_0,t;\theta)$, $\tilde{F}^{N}(x_0,y_0,t;\theta)$, $t\in [550,850]$ ms, $\theta \in \{0, \frac{\pi}{2}, \pi, \frac{3\pi}{2}\}$, are shown in Fig. \ref{Temporal-Field-x-113-y-80}. As it is shown in Fig. \ref{Temporal-Field-x-113-y-80-NoMax},  model output at direction $\frac{3\pi}{2}$, i.e., $F^N(x_0,y_0,t;\frac{3\pi}{2})$, is higher than model output at direction $0$, i.e., $F^{N}(x_0,y_0,t;0)$, during time period $[620,680]$ ms. This result conflicts with that TQD model should show the strongest response along actual motion direction, so motion direction can be inferred by determining the direction of the strongest model response. However, as we have mentioned before, for classic TQD model, incorrect signal correlation will cause confusion in motion-direction determination, especially in cluttered background. This confusion reflects in that model output along actual motion direction is not significantly higher or even lower than model outputs along other directions. Because background is in rightward motion (direction $0$), the expected result should be that $F^{N}(x_0,y_0,t;0)$ is higher than $F^{N}(x_0,y_0,t;\frac{3\pi}{2})$ during time period $[550,850]$ ms. Obviously, as we can see from Fig. \ref{Temporal-Field-x-113-y-80-NoMax}, confusion has arisen in classic TQD model outputs $F^N(x_0,y_0,t;\theta)$. Compared to Fig. \ref{Temporal-Field-x-113-y-80-NoMax}, in Fig. \ref{Temporal-Field-x-113-y-80-Max}, all model outputs at four cardinal directions of the improved TQD model, i.e., $\tilde{F}^{N}(x_0,y_0,t;\theta)$, $\theta \in \{0, \frac{\pi}{2}, \pi, \frac{3\pi}{2}\}$, are close to $0$ during time period $[620,680]$ ms. This is because $S^{ON}(x_0,y_0,t)$ and $S^{OFF}(x_0,y_0,t)$ are not local maximum in the local neighborhood $\Omega_{(x_0,y_0)}$ during time period $[620,680]$ ms. After max operation mechanism, all signals which are not local maximum in neighborhood $\Omega_{(x_0,y_0)}$, will be set as $0$. Therefore, model outputs of spatial coordinate $(x_0,y_0)$ will be close to $0$ during this time period. However, we should mention that although model outputs of the improved TQD model are $0$ during time period $[620,680]$ ms, motion direction of the background in neighborhood $\Omega_{(x_0,y_0)}$ can be inferred by local maximum in this neighborhood. This is feasible because a local maximum must exist in each local neighborhood.

In order to intuitively present detection performance of classic TQD model and the improved TQD model, model outputs of these two models corresponding to Fig. \ref{TestSet-1-Frame-840}, i.e., $F^{N}(x,y,840;\theta)$ and $\tilde{F}^{N}(x,y,840;\theta)$, are projected onto X-Y plane. Here, we should indicate that when a projection threshold $\gamma$, time $t$ and direction $\theta$ are given, spatial coordinate $(x,y)$ whose corresponding model outputs $F^{N}(x,y,t;\theta)$ and $\tilde{F}^{N}(x,y,t;\theta)$ are larger than projection threshold $\gamma$, can be shown on X-Y plane.  Projection results of Fig. \ref{TestSet-1-Frame-840} are presented in Fig. \ref{NoMax-Projection} and Fig. \ref{Max-Projection}, where projection threshold $\gamma$ is set as $0.05$. As we can see from Fig. \ref{NoMax-Projection}, classic TQD model not only show strong response along actual motion direction ($\theta = 0$), but also along other three directions. Obviously, this is not the result of what we expect. Because background is in rightward motion shown in Fig. \ref{TestSet-1-Frame-840}, TQD model should show the strongest response to actual motion direction ($\theta = 0$), but much weaker or even no responses to other directions. However, due to incorrect signal correlation mentioned before, classic TQD model may have four strong responses along four cardinal directions at a spatial coordinate. In this case, motion direction of the background cannot be obtained by determining the direction of the strongest model outputs. For this reason, confusion will arise when we determine motion direction of the background in a local region. The output of the improved TQD model is much clearer than the output of classic TQD model. In comparison with Fig. \ref{NoMax-Projection}, in Fig. \ref{Max-Projection}, the improved TQD model shows strong response to actual motion direction ($\theta = 0$), but do not respond to other directions. Therefore, we can effectively infer motion direction of the background in a local region by the direction of the strongest response of the local maximum in this local region.

In the following paper, two evaluation indexes are defined so as to quantitatively evaluate detection performance of classic TQD model and the improved TQD model. Firstly, a set of projection threshold $\{\gamma_0, \gamma_1, \gamma_2, \cdots, \gamma_n\}$ where $\gamma_0 = 0.01$ and $\gamma_{i+1}>\gamma_{i}, i \geq 0$, should be given. Then, for a projection threshold $\gamma_i$, time $t$ and direction $\theta$, we can obtain the number of points $(x,y)$ whose corresponding output $F^{N}(x,y,t;\theta)$ (or $\tilde{F}^{N}(x,y,t;\theta)$) is higher than projection threshold $\gamma_i$, denoted by $N(\gamma_i,\theta,t)$. Here, we define detection rate $DR(\gamma_i,t)$ and the normalized number of detected points $NP(\gamma_i,t)$ by the following equations, 
\begin{align}
DR(\gamma_i,t) = \frac{N(\gamma_i,\theta_0,t)}{\sum_{\theta} N(\gamma_i, \theta, t)} \\
NP(\gamma_i,t) = \frac{N(\gamma_i,\theta_0,t)}{\sum_{\gamma_i} N(\gamma_i, \theta_0, t)}
\end{align}
where $\theta_0$ is the motion direction of background.

\begin{figure}[t]
	\centering
	\subfloat[$DR(\gamma_i,840)$]{\includegraphics[width=0.2\textwidth]{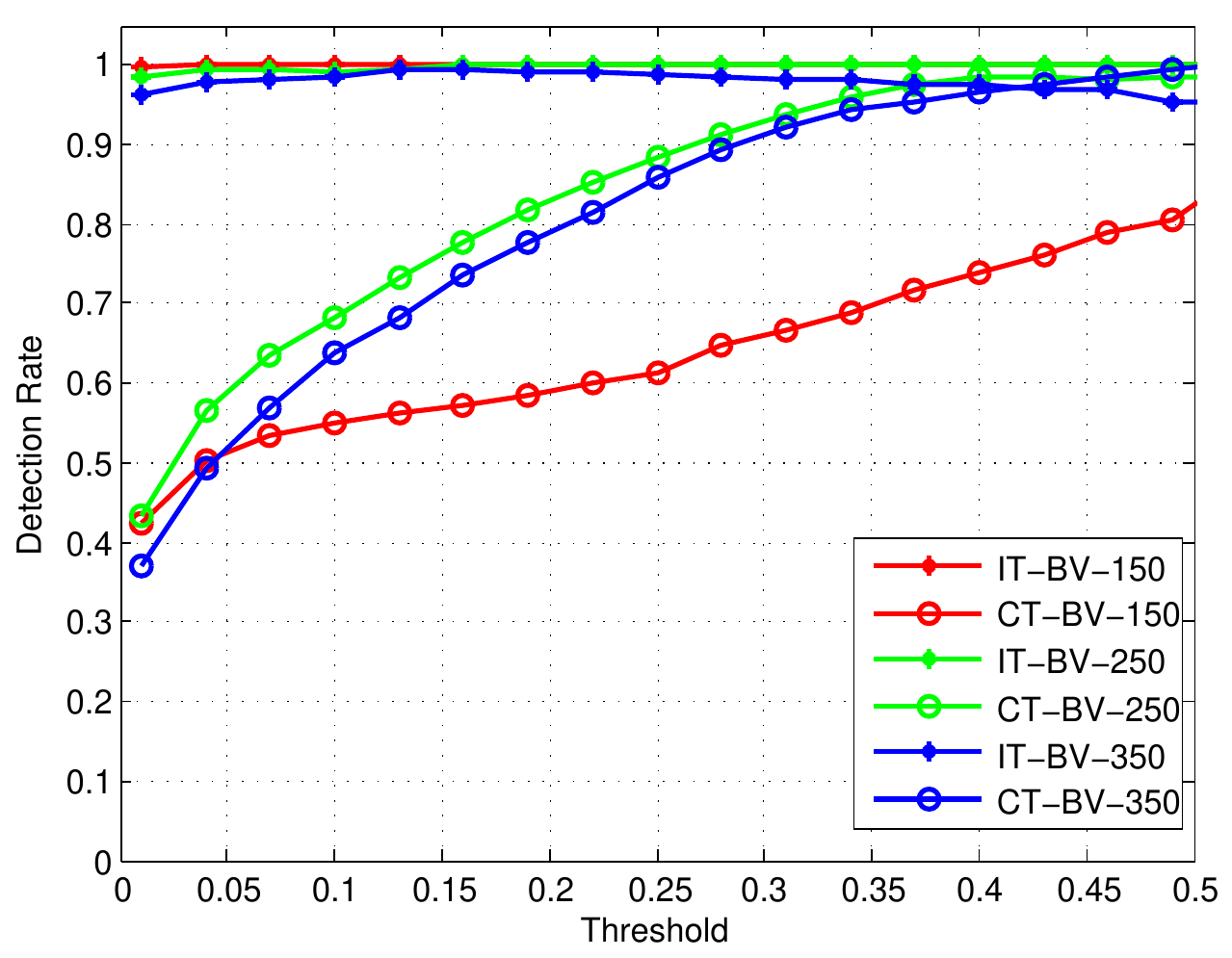}
		\label{Threshold-Detection-Rate}}
	\hfil
	\subfloat[$NP(\gamma_i,840)$]{\includegraphics[width=0.2\textwidth]{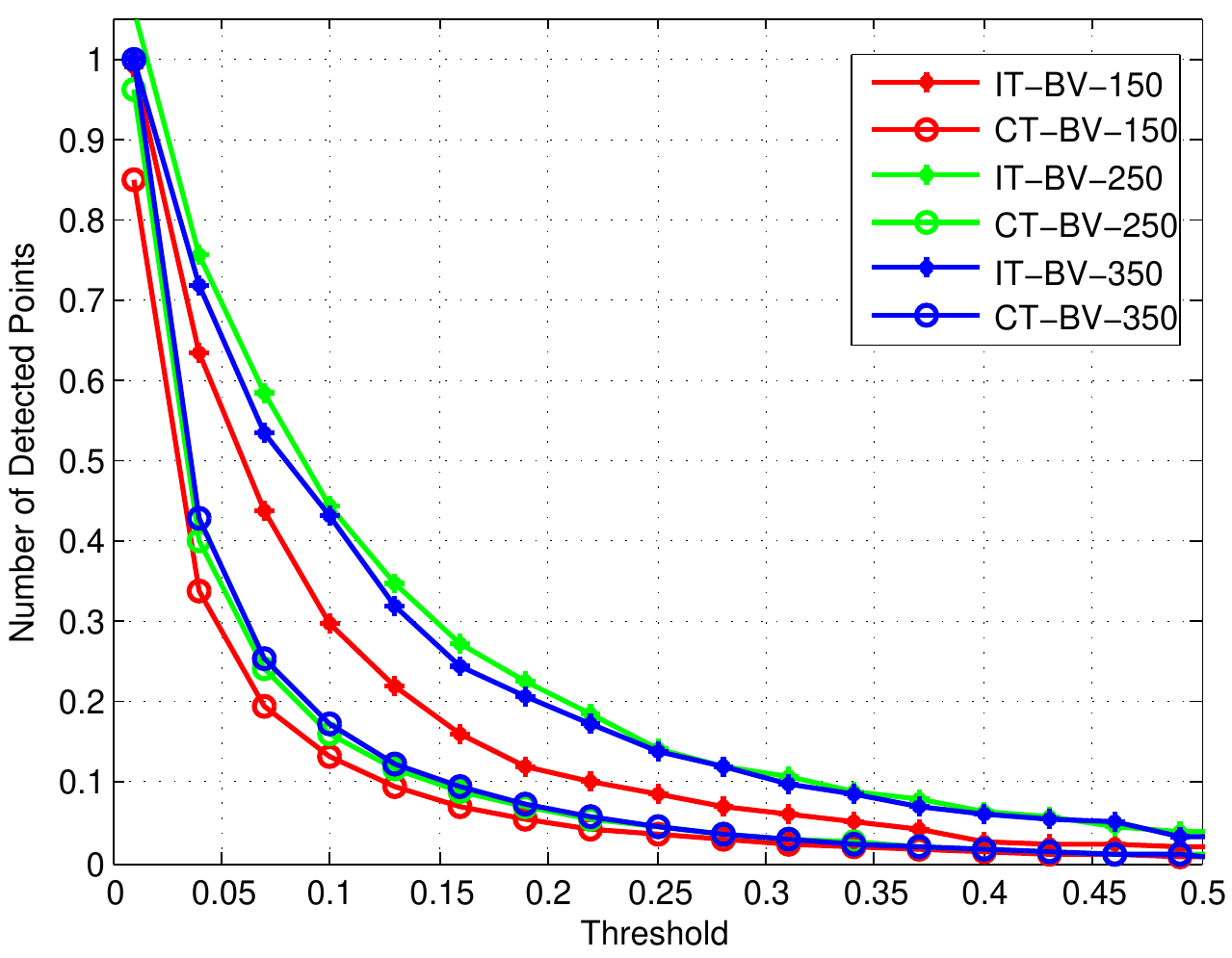}
		\label{Threshold-Number-Detected-Points}}
	\caption{Detection rate $DR(\gamma_i,840)$ and the normalized number of detected points $NP(\gamma_i,840)$ of the $840$th frame of the first image sequence. Horizontal axis denotes projection threshold while vertical axis represents Detection rate or the number of detected points. Legend IT-BV-150 and CT-BV-150 denote the result of the improved TQD model (IT) and classic TQD model (CT) when background velocity (BV) is set as $150$. Similarly for other legends.}
	\label{Result-TestSet-1}
\end{figure}

For Fig. \ref{TestSet-1-Frame-840}, i.e., $840$th frame of the first image sequence, we set background velocity as $150$, $250$, $350$ and corresponding results of classic TQD model and the improved TQD model are shown in Fig. \ref{Result-TestSet-1}. As it is shown in Fig. \ref{Result-TestSet-1}, for classic TQD model, detection rate will increase as the rise of projection threshold while the normalized number of the detected points will decrease. However, for the improved TQD model, although the normalized number of detected points will decrease as the increase of projection, detection rate shows no significant change. More precisely, detection rates of the improved TQD model are close to $1$ in despite of background and projection threshold. In general, we hope to obtain a higher detection rate at relatively low projection threshold, because the number of detected points can also reach a higher value at this time. Higher number of detected points always means that motion direction of the background can be inferred in a wider receptive field. However, as we can see from Fig.\ref{Result-TestSet-1}, detection rates of classic TQD model are much lower than that of the improved TQD model at relatively low projection threshold in despite of background velocity.

\begin{figure}[t]
	\centering
	\includegraphics[width=0.25\textwidth]{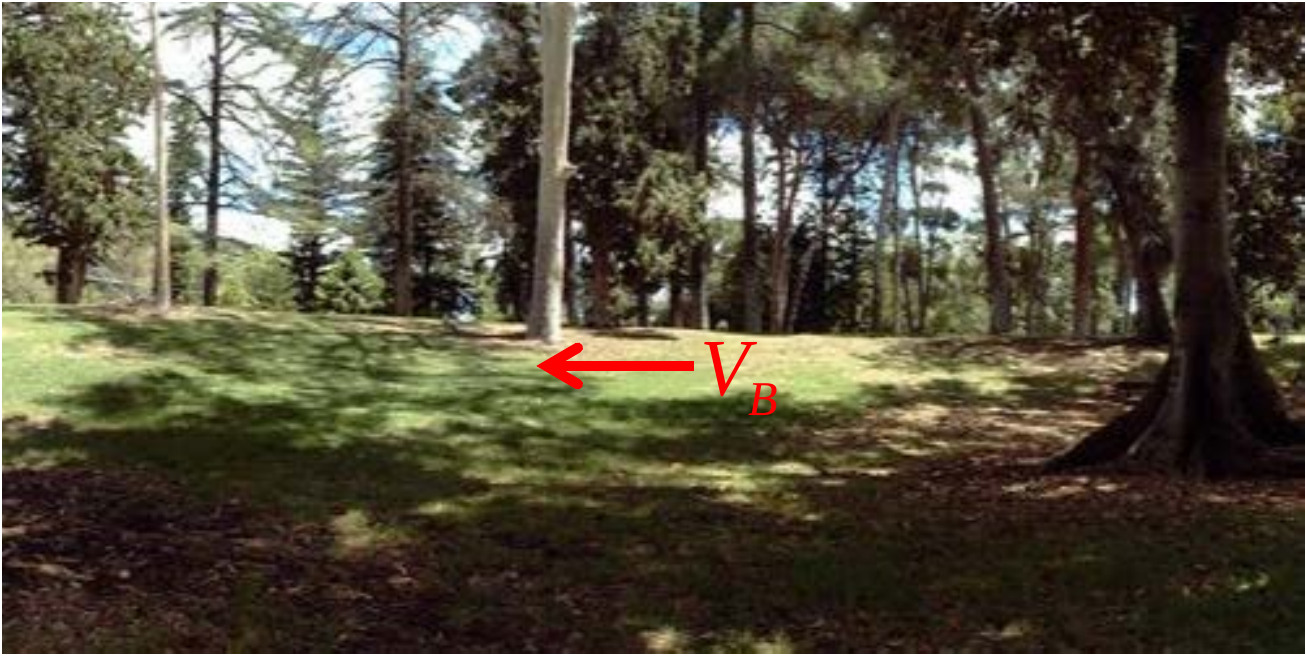}
	\caption{The $840$th frame of the second image sequence. The red arrow and $V_B$ denote motion direction and velocity of background, respectively.}
	\label{TestSet-2-Frame-840}
\end{figure}

\begin{figure}[t]
	\centering
	\subfloat[$DR(\gamma_i,840)$]{\includegraphics[width=0.2\textwidth]{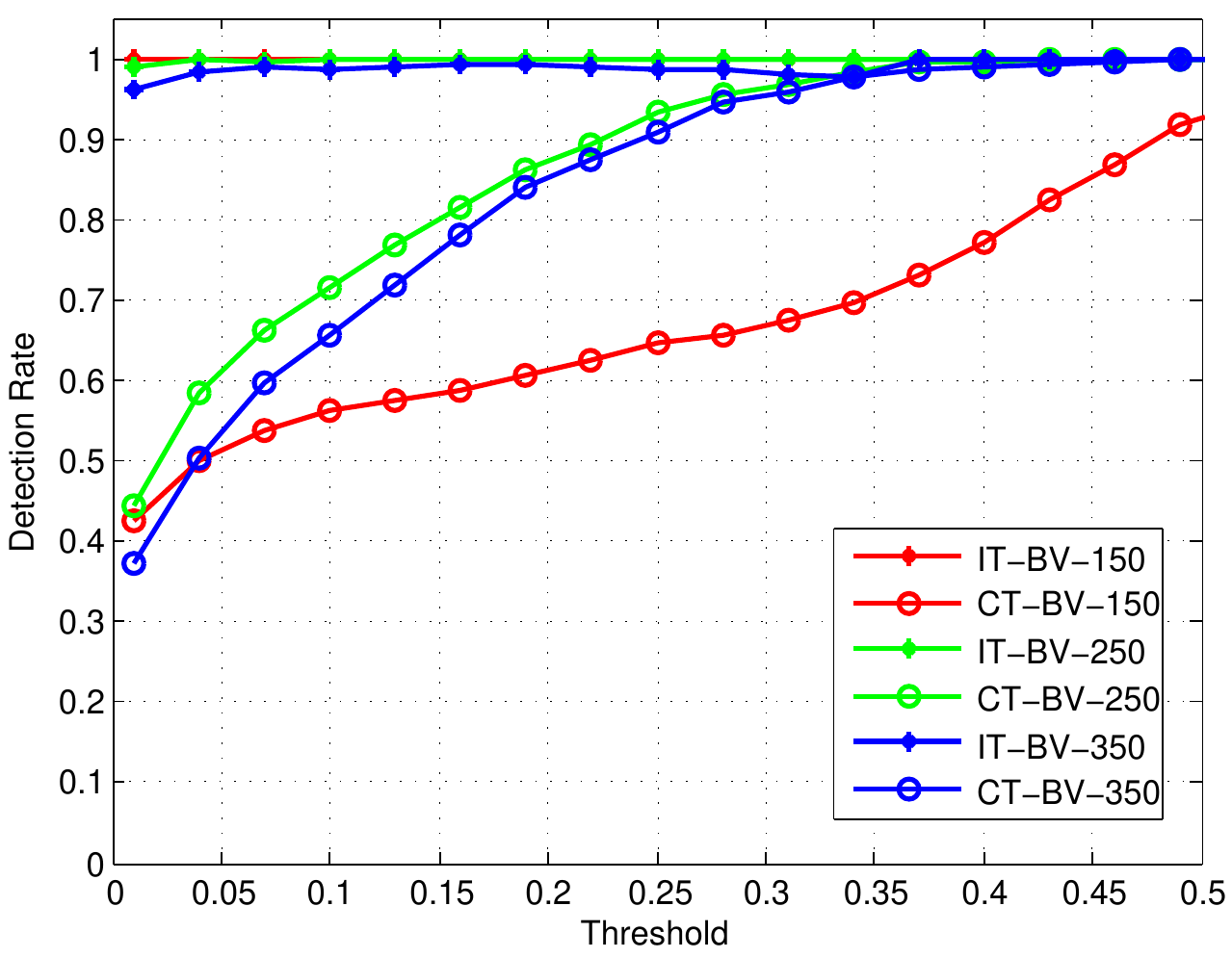}
		\label{TestSet-2-Threshold-Detection-Rate}}
	\hfil
	\subfloat[$NP(\gamma_i,840)$]{\includegraphics[width=0.2\textwidth]{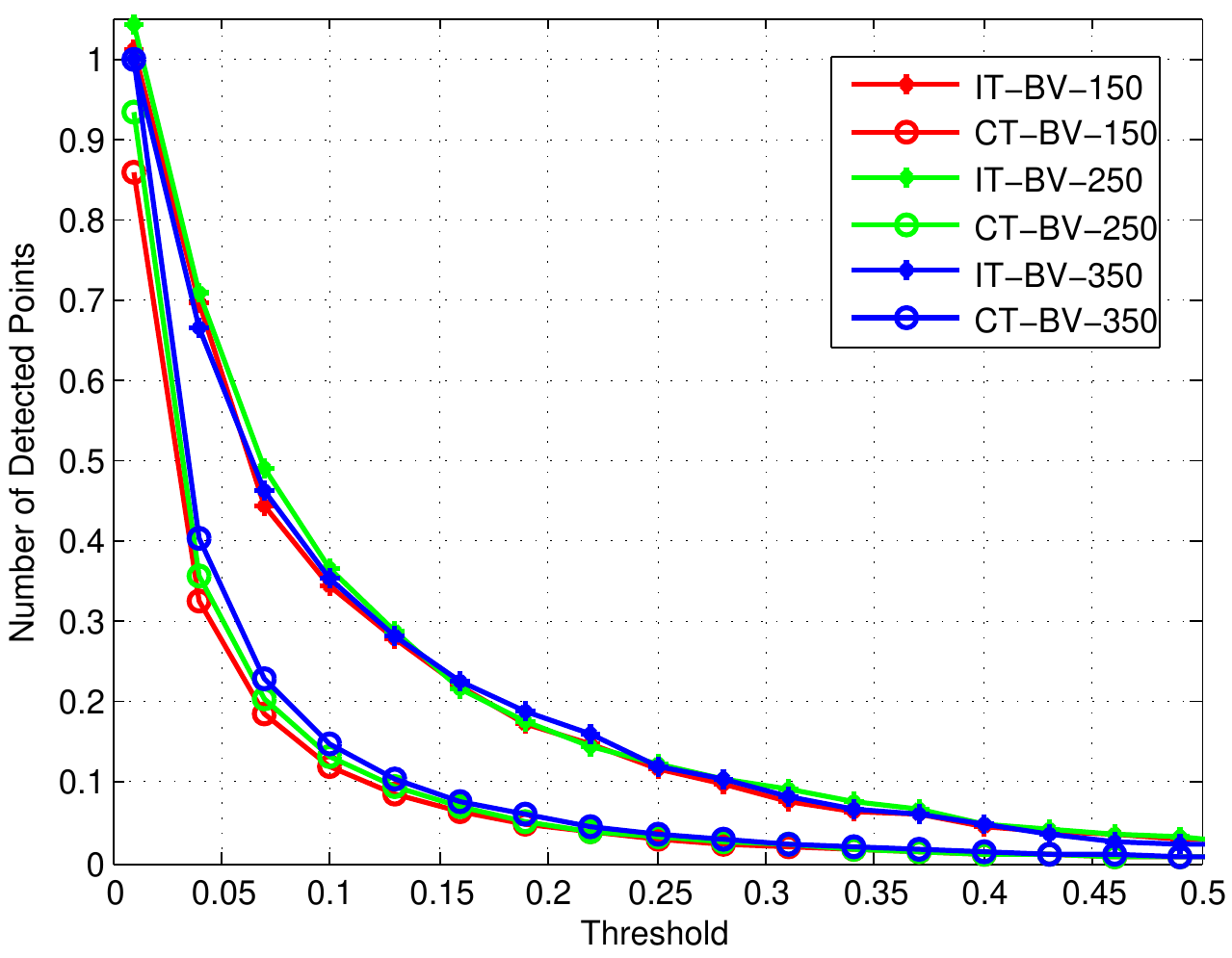}
		\label{TestSet-2-Threshold-Number-Detected-Points}}
	\caption{Detection rate $DR(\gamma_i,840)$ and the normalized number of detected points $NP(\gamma_i,840)$ of the $840$th frame of the second image sequence. }
	\label{Result-TestSet-2}
\end{figure}

\begin{figure}[t]
	\centering
	\includegraphics[width=0.25\textwidth]{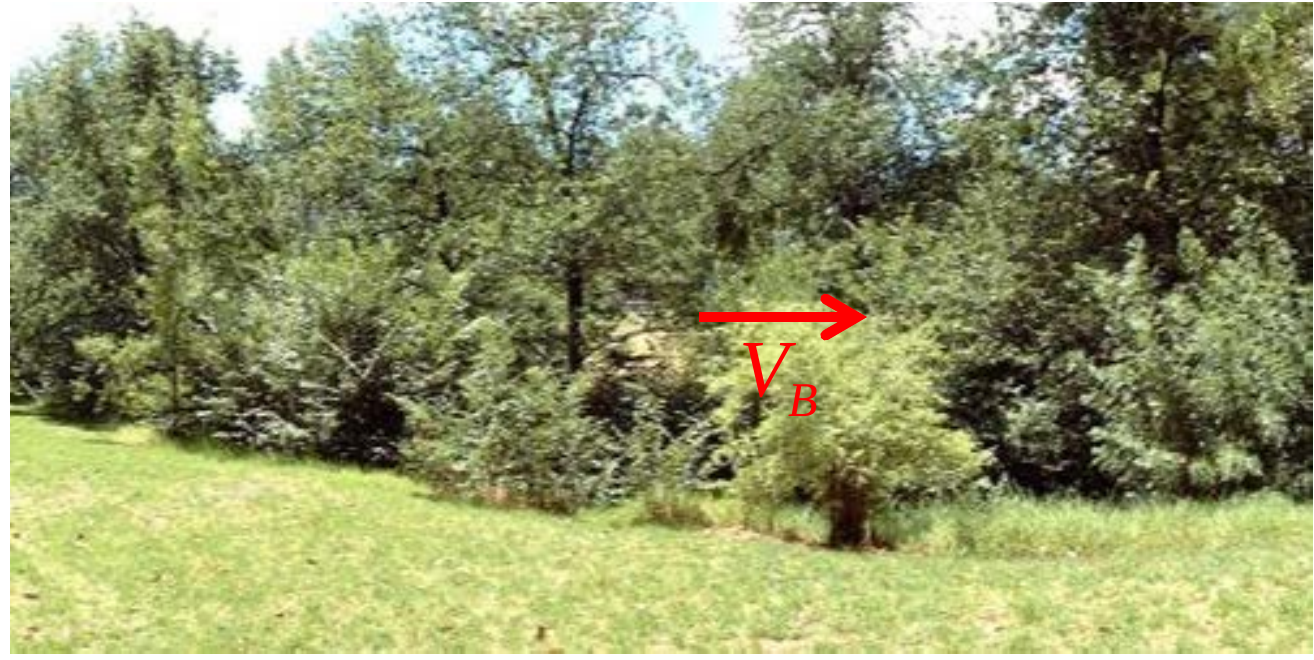}
	\caption{The $840$th frame of the third image sequence. The red arrow and $V_B$ denote motion direction and velocity of background, respectively.}
	\label{TestSet-3-Frame-840}
\end{figure}

\begin{figure}[t]
	\centering
	\subfloat[$DR(\gamma_i,840)$]{\includegraphics[width=0.2\textwidth]{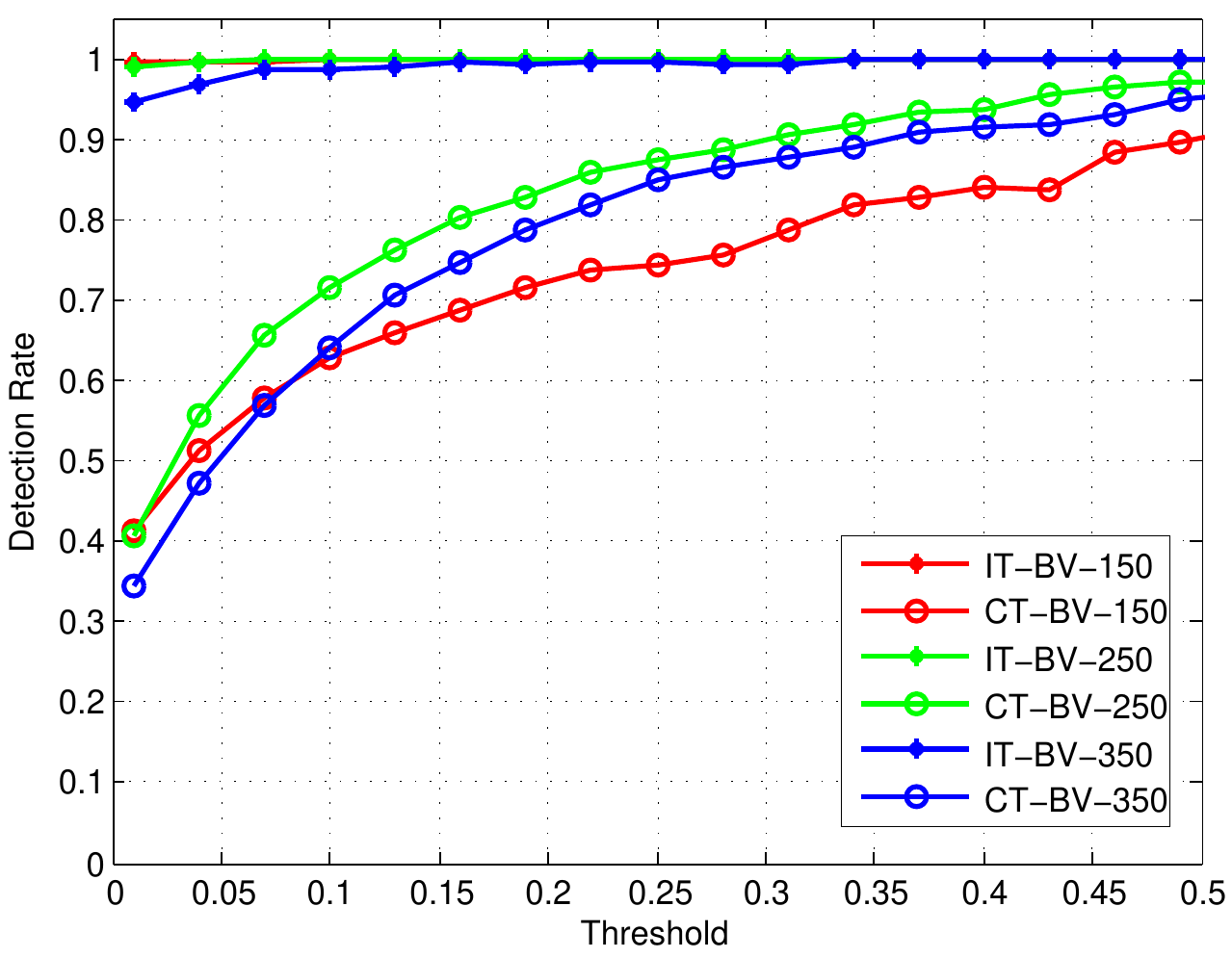}
		\label{TestSet-3-Threshold-Detection-Rate}}
	\hfil
	\subfloat[$NP(\gamma_i,840)$]{\includegraphics[width=0.2\textwidth]{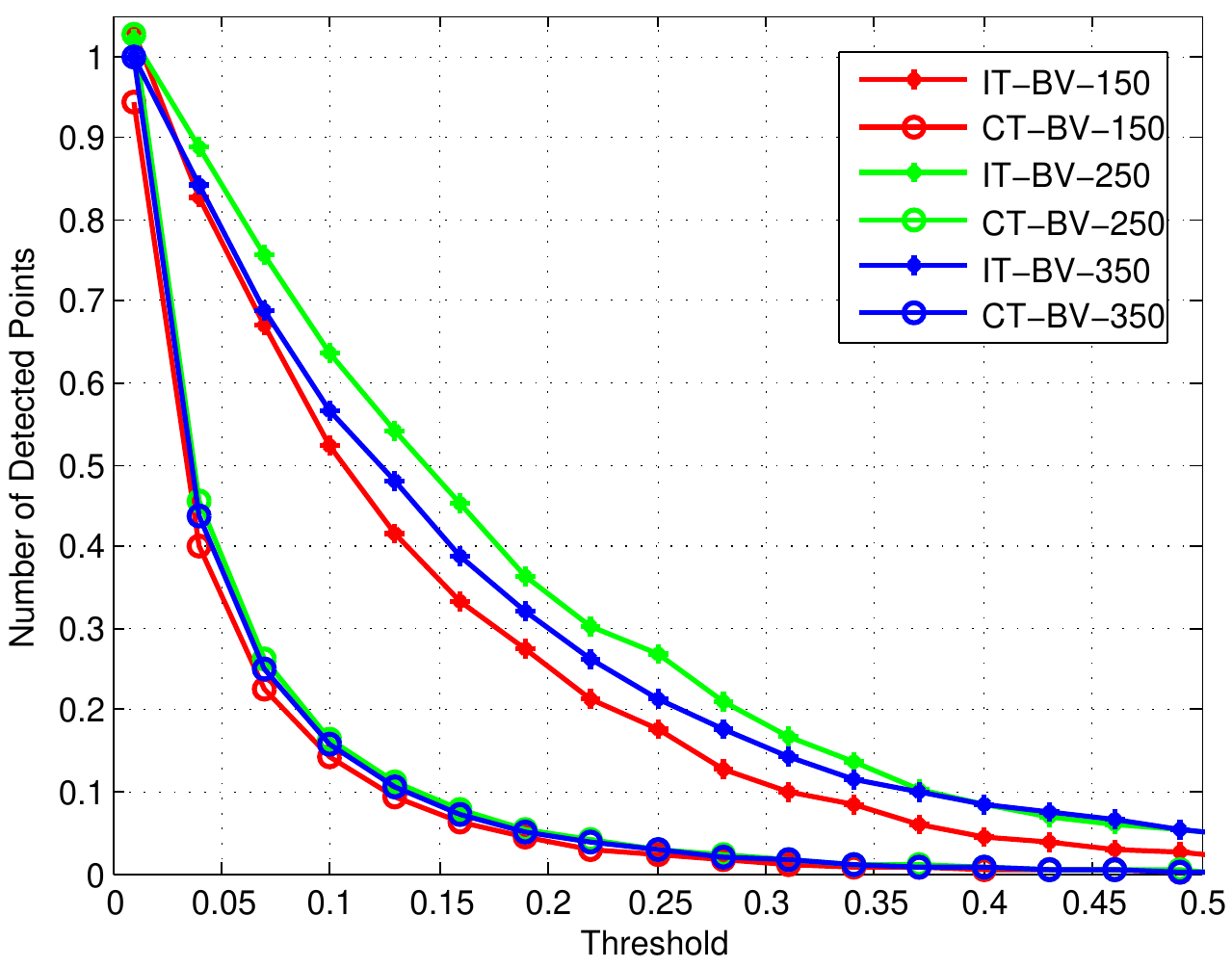}
		\label{TestSet-3-Threshold-Number-Detected-Points}}
	\caption{Detection rate $DR(\gamma_i,840)$ and the normalized number of detected points $NP(\gamma_i,840)$ of the $840$th frame of the third image sequence.}
	\label{Result-TestSet-3}
\end{figure}

The second and the third image sequences were also used to evaluate detection performance of these two models. The $840$th frame of these two image sequences are presented in Fig. \ref{TestSet-2-Frame-840} and Fig. \ref{TestSet-3-Frame-840}, respectively. In Fig. \ref{TestSet-2-Frame-840}, background is in leftward motion while in Fig. \ref{TestSet-3-Frame-840}, background is in rightward motion. Relevant results are shown in Fig. \ref{Result-TestSet-2} and Fig. \ref{Result-TestSet-3}. As we can see from Fig. \ref{Result-TestSet-2} and Fig. \ref{Result-TestSet-3}, variation trends of curves do not show significantly different from that of Fig. \ref{Result-TestSet-1}. Therefore, similar conclusion can be obtained from Fig. \ref{Result-TestSet-2} and Fig. \ref{Result-TestSet-3}. That is, optic-flow perception performance of the improved TQD model is much better than that of classic TQD model.

\section{Conclusion}
In this paper, a max operation mechanism is proposed to simulate physiological properties of a newly-identified intermediate neuron, Tm9. The functionality of Tm9 neuron was not reflected in previous correlation models, such as EMD and TQD. This max operation mechanism which acts on ON and OFF signals after signal rectification, is able to improve detection performance of classic TQD model in wide-field motion perception. Synthetic visual stimuli experiments demonstrate that this max operation mechanism can help TQD model avoid confusion in model outputs caused by incorrect signal correlation.

\section*{Acknowledgment}
This research was supported by EU FP7-IRSES Project
EYE2E (269118), LIVCODE (295151), HAZCEPT (318907), HORIZON project STEP2DYNA (691154) and ENRICHME (643691).

\bibliographystyle{IEEEtran}

\bibliography{Conference-Paper}

\end{document}